\documentclass{article}
\usepackage[T1]{fontenc} 
\usepackage[utf8]{inputenc} 
\usepackage{ismir,amsmath,cite,url}
\usepackage{graphicx}
\usepackage{color}
\usepackage[bookmarks=false]{hyperref}

\definecolor{gyomnotecolor}{RGB}{153, 0, 76} 
\definecolor{maxnotecolor}{RGB}{200, 0, 0}
\newcommand{\gyomnote}[1]{\textcolor{gyomnotecolor}{(G : #1)}}
\newcommand{\reminote}[1]{\textcolor{gyomnotecolor}{(R : #1)}}
\newcommand{\maxnote}[1]{\textcolor{maxnotecolor}{(M : #1)}}

\renewcommand{\gyomnote}[1]{}
\renewcommand{\reminote}[1]{}
\renewcommand{\maxnote}[1]{}

\usepackage{amssymb,amsthm}

\usepackage{subfiles}

\usepackage{lineno}

\title{DeepDrummer : Generating Drum Loops \\
using Deep Learning and a Human in the Loop}

\fourauthors
  {Guillaume Alain (*)} {Mila}
  {Maxime Chevalier-Boisvert (*)} {Mila}
  {Fr\'ed\'eric Osterrath} {Mila}
  {R\'emi Pich\'e-Taillefer} {Mila}

\begin{document}

\maketitle

\begin{abstract}
DeepDrummer is a drum loop generation tool that uses active learning to learn the preferences (or current artistic intentions) of a human user from a small number of interactions. The principal goal of this tool is to enable an efficient exploration of new musical ideas. We train a deep neural network classifier on audio data and show how it can be used as the core component of a system that generates drum loops based on few prior beliefs as to how these loops should be structured.

\vspace{0.3em}  
We aim to build a system that can converge to meaningful results even with a limited number of interactions with the user. This property enables our method to be used from a cold start situation (no pre-existing dataset), or starting from a collection of audio samples provided by the user. In a proof of concept study with 25 participants, we empirically demonstrate that DeepDrummer is able to converge towards the preference of our subjects after a small number of interactions.
\end{abstract}


\section{Introduction}\label{sec:introduction}


Modern day music production typically relies on a variety of software tools. These tools can help make music creation more accessible by streamlining an artist's workflow. Machine learning offers the promise that we may be able to build tools that not only automate certain tasks, but also have a certain level of understanding of an artist's musical taste and vision. 
In this work, we address the challenge of training a deep learning model to approximate the musical tastes of a human user with limited feedback for the purpose of creating an interactive tool that generates drum loops.
This approach has a wide range of potential applications, one of which is to help creators find inspiration.

\vspace{0.3em}  
We present a system that we call DeepDrummer, which is composed of three main components: the interface, the critic and the generator. Users are faced with an interface where they sequentially listen to individual drum loops and rate them by clicking on either \emph{like} or \emph{dislike} buttons.
For each user, we train a deep neural network from scratch to predict their ratings based on audio signal.
This neural network constitutes the \emph{critic} that judges the quality of drum loops produced by the \emph{generator}. 
The generator is a function that outputs random grid sequencer patterns with 16 time steps during which 4 randomly-selected drum sounds can be triggered.
We choose a very basic generator that does not have any trainable parameters, and constitutes a source of patterns that has few priors on musical structure.

Combined together, the feedback from the critic can serve as a powerful filter for the output of the generator. As a result, the interface will present only the most relevant drum loops to the user for rating.

DeepDrummer learns interactively while gathering data from a limited number of human interactions.
This is somewhat contrary to the common wisdom in deep learning that training a useful deep neural network necessarily requires hundreds of thousands, or even millions of data points. \reminote{je me répète, mais l'emploi de l'hyperbole est pas formelle}

Since human preferences are subjective by nature and may change as they are being measured, creating a perfect model of a person's musical preferences is a non-goal. We are instead interested in creating a useful filter that is utilized to quickly explore musical ideas in a way that is artistically useful.

\vspace{0.5em}  
The core contributions of this paper are the following:
\begin{itemize}
    \item We present DeepDrummer, a system that features a novel approach to combine deep learning models and active learning to generate drum loops based on limited human interactions.
    \item We run an experiment with 25 participants to empirically measure the performance of such a system, and we show that meaningful gains are made within 80 interactions.
    \item We publish DeepDrummer as an open source software, as well as all the data we collected during our experiment, which includes 3500 generated drum loops and associated user ratings.
\end{itemize}

We organise our work by first detailing the components of our interactive framework, followed by our proposed experimental protocol. Finally, we analyse our empirical results and discuss future work.

\section{Related Work}\label{sec:related-work}

The problem of music generation has attracted a much attention throughout the Music Information Retrieval (MIR) community in recent years, in part because progress in deep learning opens many interesting research directions.
Many methods frame music generation as a sequence modeling problem \cite{oord2016wavenet, manzelli2018conditioning, dhariwaljukebox, gillick2019learning, DBLP:conf/iui/VoglK16, DBLP:conf/waspaa/LattnerG19}. The general assumption is that the right probabilistic model will capture a manifold of desirable music to draw from.

Another popular approach to generate any kind of artistic content is that of Generative Adversarial Networks (GAN) \cite{goodfellow2014gans}. The core idea is that a distribution can be learned by combining two neural networks, a generator and a discriminator, in a configuration that represents the equilibrium of a game where the two networks are trying to achieve opposite goals. This differs from the classic approach of loss minimization.
There have been a variety of applications to the concept in the domain in music generation \cite{dong2018musegan, engel2019gansynth, donahue2018adversarial, kumar2019melgan, vogl2018deep}.
DeepDrummer draws some inspiration from the concept of generator and discriminator networks of GANs, but there is no adversarial training in DeepDrummer. Moreover, there is no reference dataset that the generator seeks to reproduce. We use the term \emph{critic model} to highlight the fact that this is not a GAN discriminator.

DeepDrummer fits in the trend of computer music generation by integration of deep learning in the composition process, but it does not fall in the sequence-modeling approach. Instead, one of the novel contributions of DeepDrummer is to have a generator-critic framework in which the computational heavy lifting is done exclusively by the \emph{critic} model. Our work shares similarities with the work of 
\cite{jaques2017tuning} in which an agent in a reinforcement learning environment produce music using a sequential model, but gets more rewards from an external set of rules encouraging adherence to music theory.

DeepDrummer receives feedback from users through a simple interface that involves a binary choice of \emph{like} or \emph{dislike}. This kind of interaction is reminiscent of the work of \cite{NIPS2007_3219} in which the authors use Gaussian processes in order to navigate a complex space of parameters based on minimal user feedback. Their motivating use case involves graphics rendering in which the meaning of the exposed parameters does not translate intuitively into the visual output of the system. Humans are very good at judging the visual output of such systems. It then make sense for a system to capitalize on this ability (this is also discussed in \cite{wilson2012bayesian}). Incidentally, this concept from \cite{NIPS2007_3219} has been implemented in a musical setting by \cite{10.1145/2557500.2557544}, in which the authors developed a platform that applies active learning to learn higher-level intuitive synthesizer knobs by querying users about perceived sound quality.
One of the notable differences in the case of DeepDrummer compared with the work of \cite{NIPS2007_3219, 10.1145/2557500.2557544} is that we are not solving an optimization problem with the goal of reaching the global maximum. Our goal is not the find the one specific drum loop that the user might have in mind, but rather to suggest many good drum loops that fall in the same family as what the user rated positively.




In active learning, a system queries the user strategically, so as to gain as much useful information as possible while minimizing the number of interactions. In a context where a classifier is trained, the system will often \hbox{focus} on the decision boundary where all the ambiguous data points are found. In the case of DeepDrummer, a similar phenomenon occurs, though maximal learning opportunities instead arise when DeepDrummer produces erroneous drum loops that it confidently feels that the user will \emph{like}. Much like in the case of active learning, the training set grows over time, but contains only data points that are highly informative in order to correct misconceptions the critic has about the user.
See \cite{christiano2017deep} for a recent example of reinforcement learning with minimal human interactions.

Finally, there is also an element of commonality between DeepDrummer and music recommendation systems (MRS) that are based on audio similarity between songs \cite{bogdanov2011content, lops2011content, wang2014improving, zangerle2018content}. DeepDrummer's critic model takes audio data in order to make a prediction of the user's probability of \emph{liking} it. This allows DeepDrummer to generalize across the various sounds found in drum loops.

\section{Model and Framework}\label{sec:nnetaudioclassifier}

\subsection{Overview}\label{sec:description}

The basic pipeline for DeepDrummer consists of the following components:
\begin{itemize}
    \item a random drum pattern \emph{generator} to propose initial grid sequencer patterns;
    \item a library of one-shot audio samples including the kind of sounds usually found in a drum kit;
    \item a function that renders grid sequencer patterns and a list of associated drum samples (one per grid row) into an audio waveform;
    \item a neural network classifier \emph{critic} to determine the desirability of the drum loop audio by outputting a value in the interval $[0, 1]$, which is a prediction of the odds of the user \emph{liking} that drum loop;
    \item a web interface to present the human user with drum loops and get feedback as \emph{like}/\emph{dislike} ratings.
\end{itemize}

\begin{figure}[htb!]
 \centerline{
 \includegraphics[width=\columnwidth]{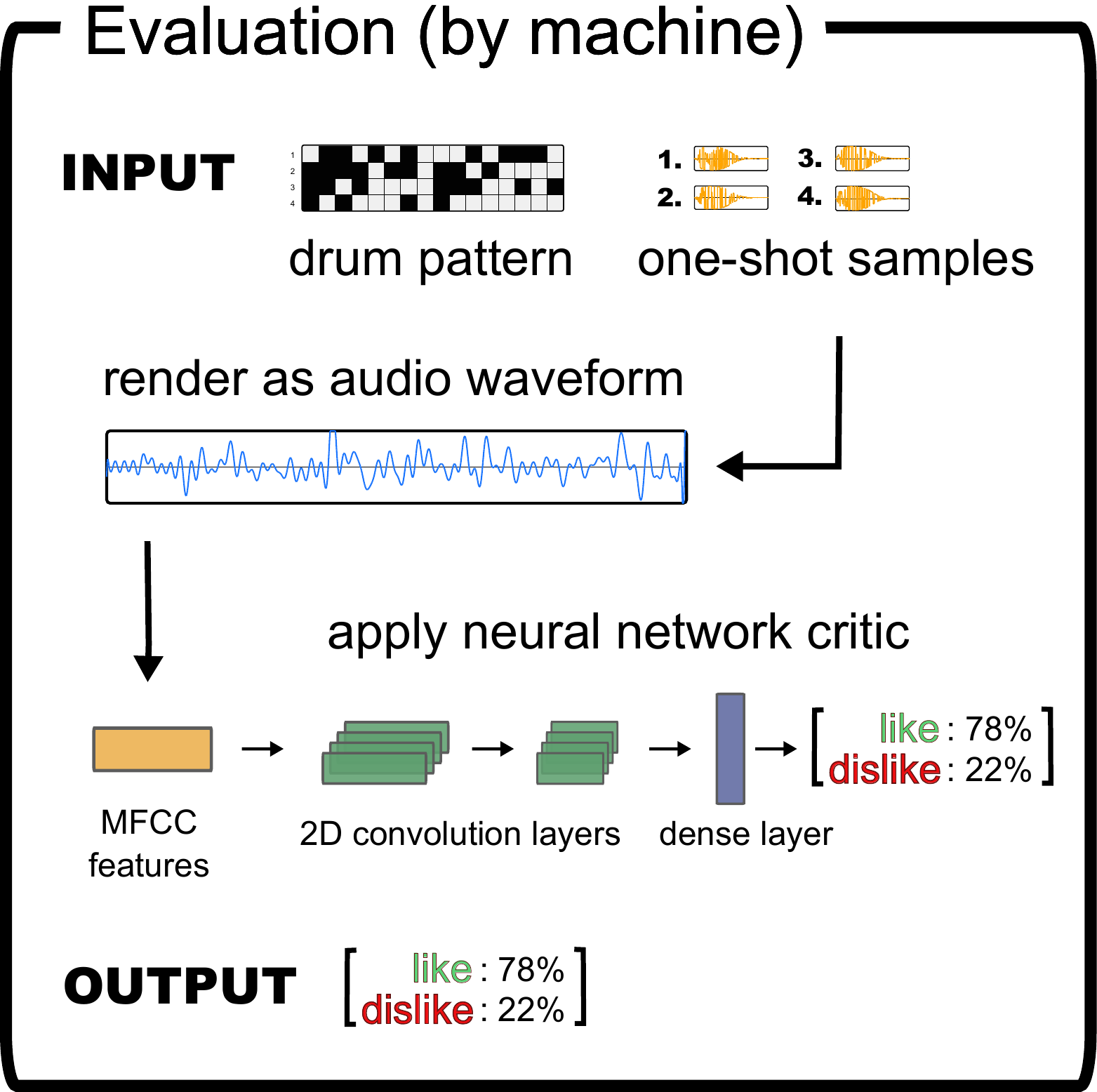}}
 \caption{Sketch of the core components of DeepDrummer. Drum patterns are generated at random, with one-shot samples (e.g. kick, snares, hi-hats) assigned to each track, and then rendered as audio waveforms. A neural network \emph{critic} takes that drum loop audio as input (as MFCC features) and makes predictions about whether the current user will \emph{like} or \emph{dislike} it. We can search through the space of drum patterns, through incremental perturbations, to gravitate towards drum loops which we believe the user is more likely to \emph{like}. We explored alternatives to MFCC features and found that MFCC yielded better performances empirically in our limited-data regime.}
 \label{fig:evaluation-by-machine}
\end{figure}

We refer to \figref{fig:evaluation-by-machine} for a visual depiction of the core steps of this pipeline where the critic evaluates the potential of a drum pattern and instrument assignment.

In this paper, we use the concept of ``drum loop'' to refer simultaneously to the following two alternative representations. 
When talking about perturbations or hill climbing on a drum loop, we mean the symbolic representations comprising a drum pattern as well as assigned instruments (i.e. one-shot samples). When we talk about applying a neural network (i.e. the critic) to a drum loop, we always mean the audio waveform and not the symbolic representation. Note that both of those representations are illustrated in \figref{fig:evaluation-by-machine}.

In the context of the experiments described in this paper, for simplicity, we are always working with 16-step patterns with 4 instrument tracks (4 one-shot samples). These form one bar, or two seconds of audio at 120bpm, which is rendered into monophonic audio at 44.1kHz. We draw from a varied collection of 340 one-shot samples (see Appendix \ref{app:samples}).

\subsection{Network Architecture of the Critic}\label{sec:network-architecture-critic}

A few different network architectures were considered for the critic (see Appendix \ref{sec:alternatives}), but the architecture we ultimately chose is one that uses Mel-Frequency Cepstral Coefficients (MFCCs) \cite{davis1980mfccs} as input features. This choice was made because DeepDrummer operates in a regime with very limited training data, making it difficult to train 1D convolutions to learn input features from raw audio.

The network uses a stack of 4 layers of 2D convolutions (with 64, 64, 64, then 8 channels), each with leaky ReLU activation and batch normalization to downsample the MFCC features. The dimensions of the hidden representations is reduced each time by using convolution kernels of size $(4, 8)$ and strides $(2, 4)$.

This is then followed by one dense layer of 128 units (with leaky ReLU). A final mapping down to two units culminates in a softmax layer that estimates the probability that the user will \emph{like} or \emph{dislike} a given input. The network is relatively small at just 291K learnable parameters. This is intentional, as small networks tend to be easier to train with limited data. Dropout and weight decay regularization are used to prevent overfitting. For more details, source code is available on GitHub \footnote{\href{https://github.com/mila-iqia/DeepDrummer}{https://github.com/mila-iqia/DeepDrummer}}.

The neural network is trained in a supervised fashion based on the feedback given by the user. The fact that it takes an audio input, instead of symbols, means that it can learn about what our user wants from the perspective of the sound itself instead of just from the mapping of the instruments. This makes it possible for the classifier to generalize between variants of kicks (or hi-hats, or any other sounds).

We did not manually sort instruments into categories, nor give DeepDrummer any idea that certain sounds should play a specific role in drum loops. An upside of this approach is that it can lead to creative use of the sounds in the collection, and it is easy to add to the existing collection by simply copying audio files to a single directory.

\subsection{Training the Critic}
\label{sec:training-the-critic}
The critic $f_\omega(x) \in [0,1]$ is initialized randomly for every user, where $\omega$ are the user-specific parameters and $x$ a drum loop. The training set is constituted of all the previous drum loops rated by the user, and it grows every time we get a new rating. 
By minimizing the cross-entropy loss of the critic during training, a model that generalizes well should be such that its output $f_\omega(x)$ matches the probability that the user rate the drum loop $x$ positively, that is, 
\begin{equation}
    f_\omega(x) \approx P\left( \textrm{user rating = \emph{like}} \hspace{0.2cm} | \hspace{0.2cm} \textrm{input = } x \right)\label{eqn:mcmc_fx}
\end{equation}

The training involves frequent retraining of the critic as more drum loops are being labeled by the user, and updating the parameters $\omega$ through gradient descent minimization of the loss. The complete experimental procedure is described in more details in Section \ref{sec:experimental-protocol}. 

Note that the generator does not have trainable parameters and therefore has no use for gradient coming from the critic.

\subsection{Sampling Drum Loops}\label{sec:sampling-drum-loops}

When we trained the critic model in Section \ref{sec:training-the-critic}, we sought to find a good value for the parameters $\omega$ of our critic  $f\omega$.

In this section, we are instead looking for drum loops $x$ that will maximize the value of $f\omega(x)$ given fixed parameters $\omega$. The reader may skip this section and simply assume that we have a way to sample drum loops $x$ such that the odds of drawing $x$ are proportional to $f\omega(x)$. That is, the better drum loops occur more often.  
\vspace{1em}

We start with a uniformly-random grid pattern as well as a random selection of audio samples, and then apply a sequence of perturbation to improve it.

Greedy hill-climbing optimization offers a simple method of convergence to a local maximum $x^*$ after $M$ iterations. Starting from a random drum loop $x^{(0)}$, elements in the sequence are accepted only if the transition improves the current score of the critic. Although local maxima are reached quickly, there is no guarantee they represent samples of good quality, nor that they are sufficiently varied.

To address these concerns, we instead employ MCMC with the Metropolis-Hastings method to sample drum loops \cite{MacKay2002}. This method allows to sample from any distribution where the density is known up to a multiplicative constant (i.e. it does not need to be normalized). In our particular case, we get a stationary distribution  where the likelihood of any state $x$ is proportional to $f_\omega(x)$. Sampling from this distribution only relies on the capacity to quickly evaluate $f_\omega(x)$ and to apply small perturbations to samples. This method is flexible and has the ability to visit multiple modes of the probability function without getting stuck. 
This process is described in more details in Section \ref{sec:generation-metropolis-hastings} of the Appendix.

Our experiment described in Section \ref{sec:phase-descriptions} is decomposed into Phase I and Phase II. In Phase I, we sample from Equation (\ref{eqn:mcmc_fx}) directly, with the idea that this can help the diversity of patterns proposed, while still focusing on the ones that are more probable to be \emph{liked} by the user. In Phase II, we want to present the user with drum loops that are meant to reflect the true discriminating power of the critic. To that purpose, we continue the Metropolis-Hastings iterations until we find a sample $x'$ for which $f(x') \geq 0.95$. That is,
\begin{eqnarray}
    \textrm{Phase I probability of $x$} & \propto & f_\omega(x) \label{eqn:phase1p}\\
    \textrm{Phase II probability of $x$} & \propto & f_\omega(x) \mathbb{I}\left(f_\omega(x) \geq 0.95 \right) \hspace{1.5em} \label{eqn:phase2p}
\end{eqnarray}
where $\mathbb{I}(\cdot)$ is the indicator function.


\section{Experimental Protocol}
\label{sec:experimental-protocol}

\subsection{Hypothesis and Protocol}

\begin{figure}[htb!]
 \centerline{
 \includegraphics[width=\columnwidth]{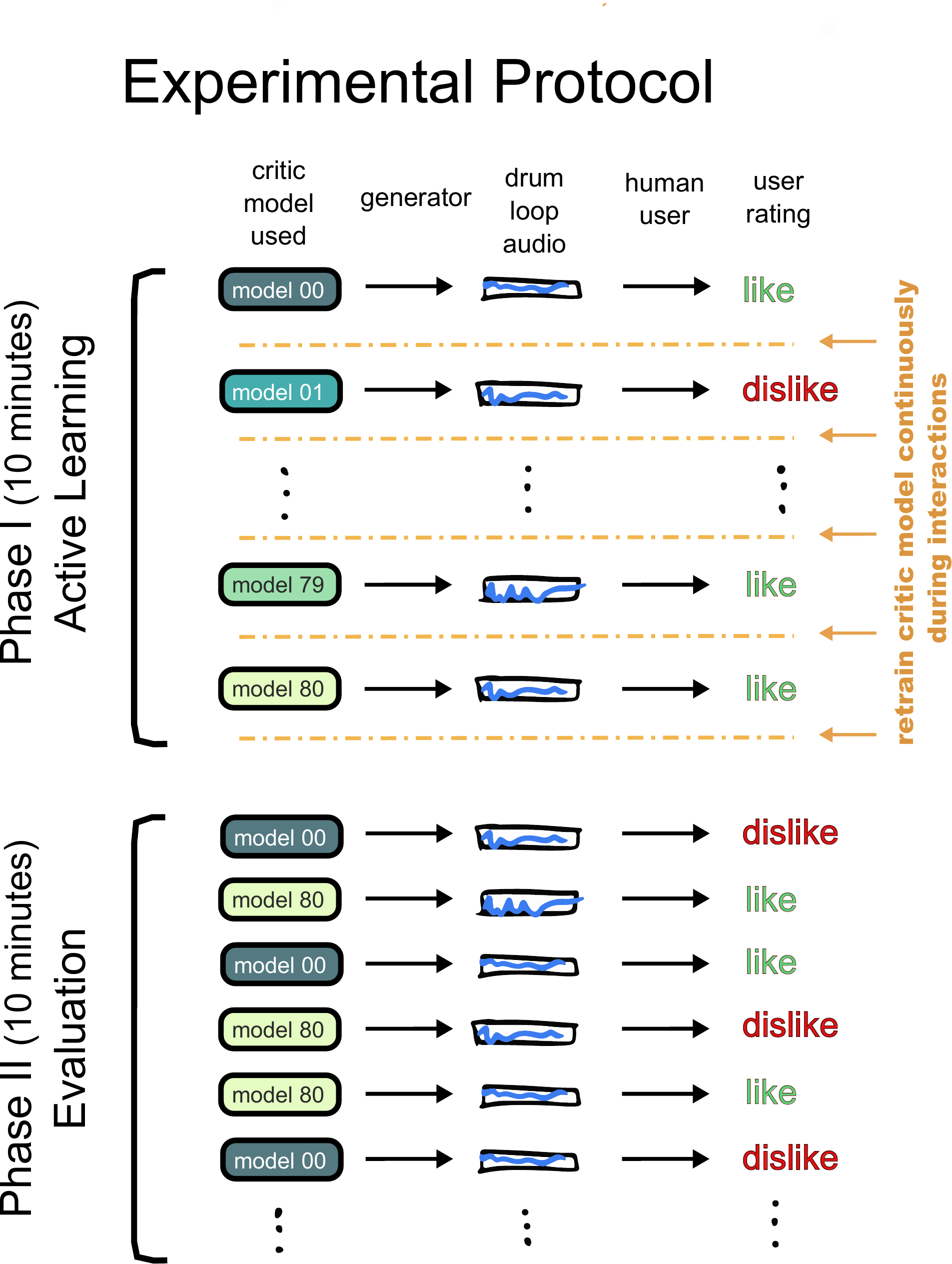}}
 \caption{We separate our experiments into two phases. The principal reason for this separation is that we need to be able to evaluate and compare fairly the performances models from different points in time.
We present the users with drum loops and ask for a binary rating \emph{like} or \emph{dislike}. The generating process uses a neural network critic to learn the tastes of the user such that only the most promising drum loops will pass through for rating.
In Phase I, we update the critic model after each rating from the user. We ask the user for 80 such ratings, and each new drum loops comes from a new updated model in the sequence.
In Phase II, we produce more drum loops using either the first critic model (untrained) or the last critic model (presumably the best we have). Note that we are \emph{not} recycling the same exact drum loops from Phase I. The ordering is shuffled so that the user is blind to the source. The ratings in Phase II are collected for analysis but no training will happen.
This whole process learns from scratch in the sense that the dataset is not seeded with drum loops and ratings coming from other users.}
\label{fig:experimental-protocol}
\end{figure}

We want to demonstrate that, after a minimal number of interactions, DeepDrummer can produce loops that the user is more likely to \emph{like}. The general assumption is that as more interactions are gathered, the critic model will converge towards an accurate approximation of the user's preferences.  

Because of practical constraints in experiment design, such as limited human attention span, we have intentionally kept the experiment short and minimalistic, and aim to demonstrate that this is something that can be accomplished in merely 10 minutes.
With that constraint in mind, we provide a proof of concept of the use of such a method as the core principle behind a more production-ready tool which would naturally involve many more components (see Section \ref{sec:production-tool} for more discussion on future directions).

Our goal, thus, is to measure a significant improvement in the proportion of drum loops being \emph{liked} by the users over the course of $80$ ratings. Given the binary nature of those ratings, this is equivalent to saying that we are only getting $80$ bits of information.

\subsection{Anticipating Preference Shifts}
\label{sec:preference-shifts}

The design of our experimental protocol needs to anticipate a potential
shift in user preferences \cite{kulesza2014structured, cartwright2016moving}. Our experimental protocol needs to be robust to the possibility that, even with a system that completely fails to learn anything, it might be possible on average for users to \emph{like} 20\% of loops at the beginning and then 35\% at the end, for example.

The issue addressed here does not refer to the randomness of the experiment, for which we can compensate by having many users and ratings. Rather, it points to the fact that a user's behaviour could be influenced by having spent the last 10 minutes rating drum loops. As such, we have designed an experiment with an active learning phase and an evaluation phase.

During the evaluation phase, users will rate drum loops generated by past models. In the evaluation phase, loops from earlier and later models are presented in a randomized order, so that participants do not know which model they came from, and so that the evaluation is robust to participants subconsciously increasing or decreasing their ratings over time as they listen to more drum loops.

\subsection{Phases I \& II of the Experiment}
\label{sec:phase-descriptions}

To adequately measure the improvement in performance over the course of interactions with the users, and counteract preference shifts as described in Section \ref{sec:preference-shifts}, we proceed in two phases as illustrated in \figref{fig:experimental-protocol}.
\vspace{0.5em}

In Phase I,
\begin{itemize}
    \item the critic is learning incrementally with every rating from the \hbox{current user};
    \item ratings are used for training the critic and not counted in the analysis afterwards;
    \item we generate 80 loops using Metropolis-Hastings as described in Equation (\ref{eqn:phase1p}), using the most recent critic model reflecting the latest ratings.
\end{itemize}

In Phase II,
\begin{itemize}
    \item the parameters of the critic are fixed;
    \item ratings are used purely for evaluation and analysis;
    \item we generate 60 loops using Metropolis-Hastings as described in Equation (\ref{eqn:phase2p}), half from the initial critic and half from the final critic;
    \item loops are presented to the user in randomized order.
\end{itemize}

\begin{figure}[htb!]
 \centerline{
\includegraphics[trim=0 625 200 0,width=0.75\columnwidth]{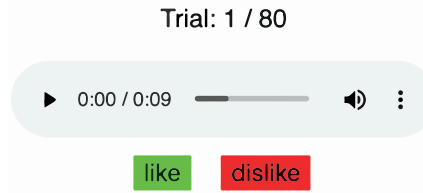}}
    \caption{This is the interface presented to the users in their web browser.
    }
\label{fig:web-like-dislike}
\end{figure}

Each drum loop generated lasts 2 seconds (one bar at 120 beats per minute) and is repeated 4 times. We also leave one more second of audio at the end for transients to fade. This is done to allow the users to evaluate the rhythmic property of each drum loop. They can, however, give their ratings before the end of the whole sequence.
We show in \figref{fig:web-like-dislike} what the main component of the interface looks like to the users.

In Phase II, we generate new drum loops that come from the critic models evaluated. We avoid recycling the same drum loops from Phase I as this could lead to specific loops being recognized. The users are given no information as to which model was used to generate the loops.

\subsection{Web-Based Experiment}

Our initial plan was to have volunteers come into a room one at a time, read a sheet of instructions, and take the experiment using a desktop computer and headphones we provided so that experimental conditions could be carefully controlled. However, the ongoing social distancing restrictions due to the COVID-19 pandemic made such a setup impossible. Instead, we opted to implement a web interface to DeepDrummer, and directed volunteers to take the experiment online. All 25 volunteers received the same set of instructions (see Appendix \ref{sec:web-splash-page}), and we refrained from giving additional information about implementation details of our project to volunteers. A video demonstrating our web interface is available on YouTube\footnote{https://youtu.be/EPKsUf5YBeM}.

\section{Results}

A total of 25 people participated in this study.
We want to determine whether the critic had a better performance at the end than at the beginning, and possibly by how much. 

Every user $i$ provides $30$ binary ratings for their initial critic model and for their final critic model. We are going to analyze the average of those ratings, such that
\begin{eqnarray*}
    \theta_\textrm{init}^{(i)} & = & \textrm{ratio of drum loops, from \textbf{initial} critic model,} \\
    & & \textrm{that user $i$ \emph{liked} during Phase II,} \\
    \theta_\textrm{final}^{(i)} & = & \textrm{ratio of drum loops, from \textbf{final} critic model} \\
    & & \textrm{that user $i$ \emph{liked} during Phase II,} \\
    \Delta\theta^{(i)} & = & \theta_\textrm{final}^{(i)} - \theta_\textrm{init}^{(i)}.
\end{eqnarray*}
It bears reminding that the critic models are not the same for any two users. By studying the distributions of those two values $(\theta_\textrm{init}^{(i)}, \theta_\textrm{final}^{(i)})$, as well as that of their difference $\Delta\theta^{(i)}$, we can get a sense of the progress made. Larger values of $\theta$ correspond to more drum loops being \emph{liked}. This $\theta$ can be interpreted as the Maximum Likelihood Estimate of the parameter of a Bernouilli distribution, representing the probability of a user to \emph{like} a drum loop.

\begin{figure}[htb!]
 \centerline{
 \includegraphics[width=\columnwidth]{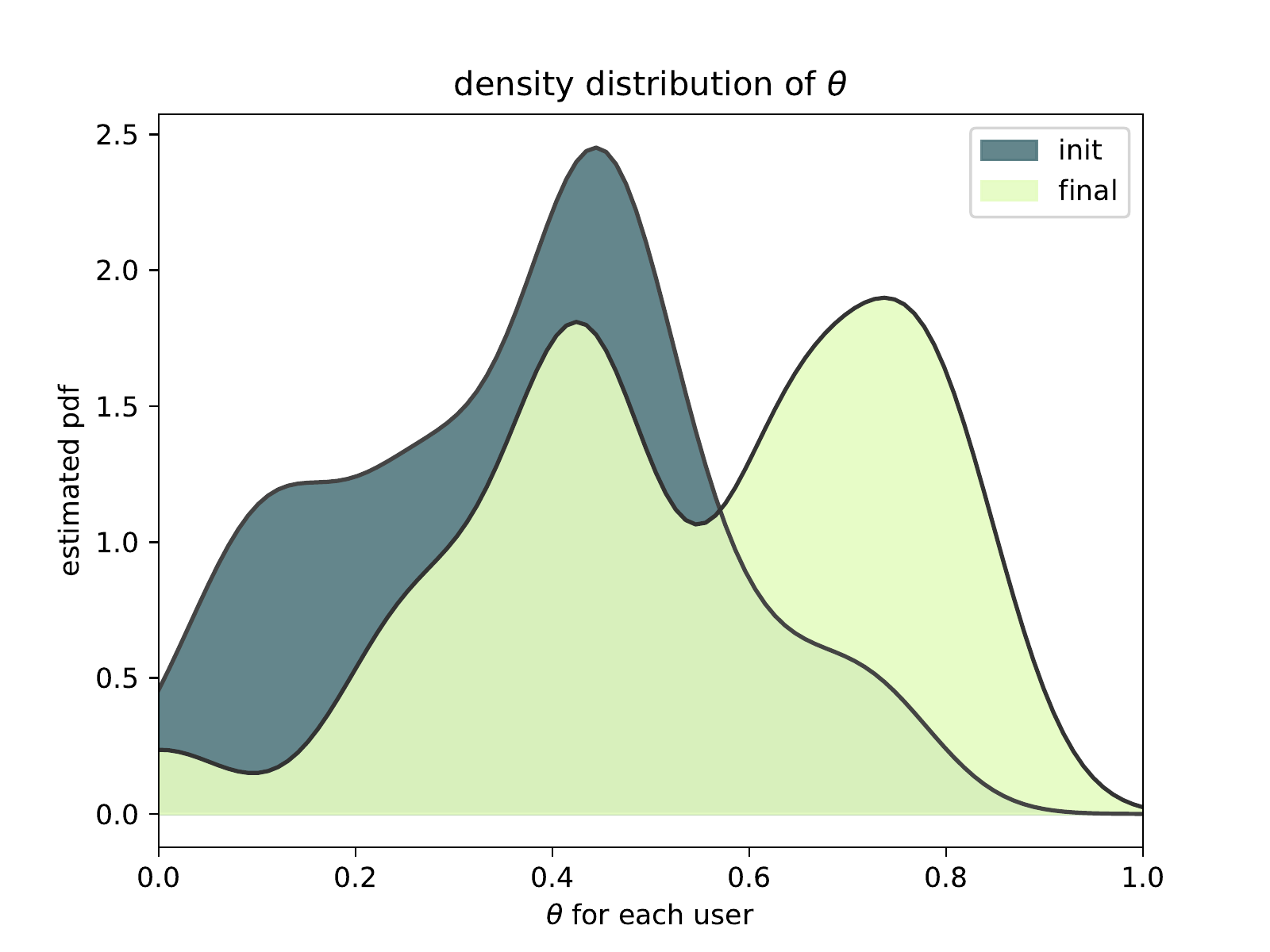}}
    \caption{This is a before/after plot. We compare the proportion of audio loops that users \emph{like} for the initial critic model versus for the final critic model, which comes after 80 \emph{like/dislike} ratings are given to DeepDrummer. We show the densities of $\theta_\textrm{init}$ and $\theta_\textrm{final}$ separately. This plot should be read as a histogram to which a smoothing kernel was applied.
    We can see that DeepDrummer without training is already capable of generating drum loops that please users $\approx 40$\% of the time. The two modes in this distribution suggests that participants tend to be almost equally split between a group who sees clear significant improvements, and another group who does not perceive much improvement (but no degradation either).
    This is the demonstration of our hypothesis. Note that those measurements are taken from Phase II of our experimental protocol, which is designed to eliminate the effects of shifting user preferences.
    }
\label{fig:results-densities}
\end{figure}

\vspace{0.3em}  
In \figref{fig:results-densities} we show the distribution of both $\theta_\textrm{init}^{(i)}$ and $\theta_\textrm{final}^{(i)}$ amongst all the users $i$. We see that the distribution of $\theta_\textrm{final}$ has more of its mass around larger values than $\theta_\textrm{init}$. Across the 25 participants, we counted that users \emph{liked} 37\% of the drum loops without any training, and with very minimal training this goes to approximately 54\% of the time. This is the demonstration of our hypothesis. We indeed have significant improvements with very few ratings (10 minutes of clicking \emph{like}/\emph{dislike}).

\begin{figure}[htb!]
 \centerline{
 \includegraphics[width=\columnwidth]{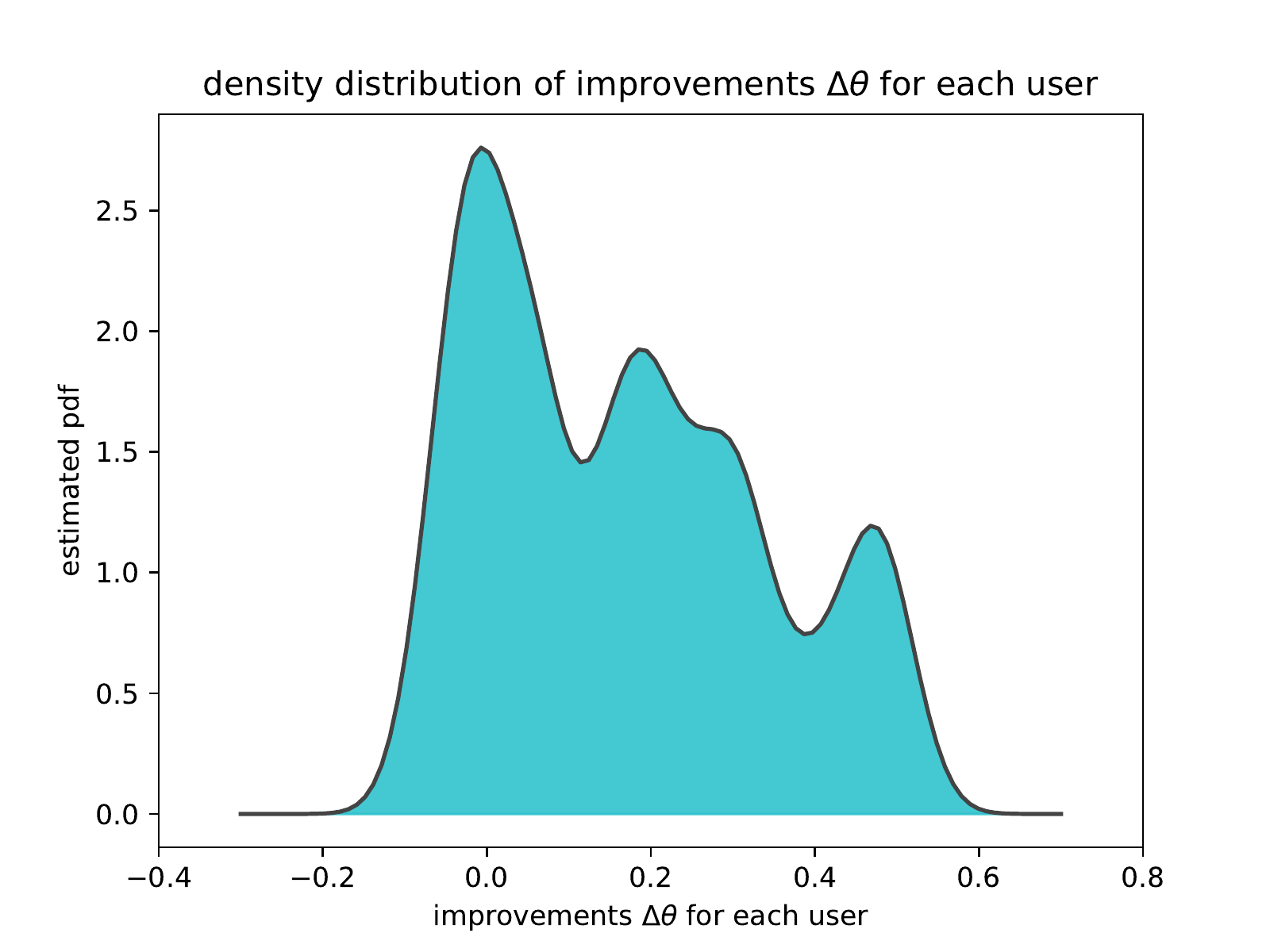}}
 \caption{We report here how much improvement was achieved during the interactions with DeepDrummer. This corresponds to the differences \mbox{$\Delta\theta^{(i)} = \theta_\textrm{final}^{(i)} - \theta_\textrm{init}^{(i)}$} among all users $i$. We can see that for the majority of users, there were clear improvements over the course of a limited number of interactions. There were $36\%$ of users with $\Delta\theta^{(i)} \geq 0.20$. This plot should be read like as histogram to which a smoothing kernel was applied.}
\label{fig:results-delta-densities}
\end{figure}

\vspace{0.3em}  
In \figref{fig:results-delta-densities}, we show the differences $\Delta\theta^{(i)}$ for each user. We can see that we have a certain percentage of users clustered around $0.0$, meaning that they saw no improvement. By combing through our data, we counted $72\%$ of users that have a $\Delta\theta^{(i)}>0$, and there was $36\%$ of users with $\Delta\theta^{(i)} \geq 0.2$.

\section{Discussion and Future Work}

\subsection{Frequencies and Rhythm}

The DeepDrummer critic learns to predict user preferences based on audio data, and is trained from a very small number of human interactions. A natural question to ask is whether the critic learns something about rhythm, or if it simply learns to filter out sounds that humans find unpleasant, such as high-frequency tones.
The experiment in Appendix~\ref{app:bestworst} partially answers this question. When trying to produce the best and
worst loops based on the critics trained for all users, we find that the worst loops are very busy, and that the loops
classified as best tend to converge towards similar rhythms. We see this as supporting evidence that DeepDrummer does learn something beyond simply filtering different sounds based on their frequency content.

\subsection{Larger Scale Experiment}

Naturally, we have wondered how well our system would perform if it were trained with more than 80 ratings and 10 minutes of human interactions. The experiment described in this paper was intentionally kept short because we did not want our
users to become tired, and for the quality of their ratings to decrease as a result. We have, however, performed many informal experiments on our own, with multiple hundreds of our own ratings. The results of such an experiment are shown in Appendix~\ref{app:supplemental}. Qualitatively speaking, it seems clear to us that DeepDrummer does yield higher quality drum loops with more training data, with much fewer false positives being produced. Given more resources, it would be very interesting to attempt to collect more training data for a Phase II experiment with many human users. Another possibility would be to begin the experiment with a system that was already pretrained on some amount of data supplied by us, rather than training DeepDrummer from scratch for each user.

\subsection{Melodies}

While we use drum loops for our experiments, in theory this could be applied to any situation in which the pattern generator supports mutations and where a hill-climbing approach would make sense. 
It seems more plausible that good drum loops would have an easier times surviving random perturbations in the space of pitch/timing (or with notes being deleted), than melodies would.
We believe that the core idea behind DeepDrummer could be used for generating music with melodies, but it may require the use of autoencoders such as in the Melody RNN work by Magenta (\cite{melodyrnn2016}). By performing the mutations in the hidden space of representations, this would increase the chances that mutations to a good melody would lead to another good melody.


\subsection{Path Towards a Production Tool}
\label{sec:production-tool}

In this paper, we focused on one particular aspect of drum loop generation. 
We present a collection of natural future directions that could be taken to bring DeepDrummer closer to a production tool:
\begin{itemize}
    \item add prior information on timing of popular drum styles;
    \item seed the training set with quality drum datasets, as though the user had an implicit liking for those drum loops before even interacting with DeepDrummer;
    \item analyze user preferences a priori, place them into clusters, and make recommendations in order to quickly gravitate towards the pre-existing clusters of drum loops to which they presumably belong;
    \item organize one-shot samples in our instrument library by type (e.g. kick, hi-hit, snare) and make
    that part of the prior on pattern generation;
    \item eliminate one-shot samples that users generally find unpleasant-sounding;
    \item allow users to input music from their own collection to seed DeepDrummer;
\item frame this as a Meta-Learning problem so as to pre-train a critic model that adapts maximally rapidly to new environments, or users in our case (see  \cite{finn2017model}).
\end{itemize}

\section{Conclusion}

In this work, we have introduced a novel way of generating music using deep learning with a human in the loop. We presented an implementation of that idea that we call DeepDrummer, which we have shown to very quickly achieve meaningful improvements by proposing drum loops to a user and receiving feedback in the form of \emph{like} / \emph{dislike} ratings.

The simplicity of the idea revolves around the use of a neural network critic that serves as a proxy for the user in order to explore a vast landscape of music very rapidly. Only the most relevant candidates are forwarded to the user.

We ran an experiment with 25 participants in which each user rated 80 drum loops. We have measured empirically the improvement of ratings after those interactions, thus confirming our claims. 

\section{Acknowledgments}
The authors would like to thank Devine Lu Linvega for many enlightening conversations, whether about the world of tools for writing music or the universe in general. We thank Anna Huang for her encouragements and guidance in framing our work as a plausible research project for publication. This research was enabled in part by support provided by Calcul Qu\'eb\'ec and Compute Canada.

\bibliography{deepdrummer}

\clearpage
\appendix

\section{Supplemental Materials}\label{app:supplemental}

\textbf{DeepDrummer : Generating Drum Loops using Deep Learning and a Human in the Loop}
\vskip 0.6em

The supplemental materials are available through a shared Google Drive link\footnote{https://bit.ly/363cBdj}. The total size of the download is 2.5GB.

\subsection{Demo Video}

A short video demonstrating the DeepDrummer web interface used for our experiments as well as the DeepDrummer desktop application trained with more samples is included in the supplemental materials. We recommend watching it, as this can give a better sense of how the software operates. This video is also available on YouTube\footnote{https://youtu.be/EPKsUf5YBeM}.

\subsection{Source Code}

The source code is open source and available on GitHub:
\href{https://github.com/mila-iqia/DeepDrummer}{https://github.com/mila-iqia/DeepDrummer}

\subsection{Drum Samples}\label{app:samples}

The 340 samples used to produce drum loops are included, in 44.1kHz mono 16-bit PCM wave format. All samples are released under the CC0 license (royalty-free, no attribution required). The samples include samples taken from famous drum machines such as the Roland TR-808, Vermona DRM1, Moog DFAM, BOSS DR-110, and also some synthesized drum hits. We intentionally included a few samples that could be described as unpleasant in order to demonstrate that a trained DeepDrummer model is able to understand that these samples are less desirable.

\subsection{Collected Data}

The data collected from all 25 users of the experiment is included. This includes all 3500 drum loops generated in phases I and II, the ratings given by each users, as well as the untrained and trained critic models. The ratings data is in a JSON format that is easy to parse. The data has been anonymized to protect the privacy of the participants.

\subsection{Best and Worst Loops}\label{app:bestworst}

We have ranked all the drum loops produced for all users in phases I and II of the experiment based on an ensemble of all trained models. This is based on an average of the score given by all models to all clips. The best 5 and worst 5 clips are included in the supplemental material. This helps illustrate what all models agree on in terms of what makes a drum loop likeable or not, and also showcases that multiple weak models can be combined together in useful ways.

The worst ranked patterns are interesting because of the contrast they provide. They serve to illustrate that DeepDrummer has learned some level of understanding of what is unpleasant to human listeners. Qualitatively, these patterns could be described as busy, and all of them contain high-frequency noises. The best ranked patterns represent what all the models agree on in terms of what is likely to be appreciated by human users. These have a clear rhythmic structure, and use drum sounds in a way that is more conventional.

\section{Alternatives Considered}\label{sec:alternatives}

Multiple alternatives were considered during the design and development of DeepDrummer. We do not have quantitative data to empirically validate each of these choices, but we have thought that it may be useful to discuss these so as to provide some additional context to our work.

\subsection{A vs B Comparison}

The current version of DeepDrummer asks the user to \emph{like} and \emph{dislike} audio clips. This paradigm can be confusing when new users are first faced with this interface, because they will naturally want to rate audio clips in relation to one another. That is, when the system is untrained, and the output is largely unappealing, a user may \emph{like} an audio clip because, even though it is not great, it is better than what they have heard so far. As such, it may seem natural to want an interface where, instead, users are presented with two audio clips and asked to pick their favorite. We have tried this idea in an early version of DeepDrummer, and found it problematic, because it seems that, when comparing two clips $A$ and $B$, users tend to forget what $A$ sounded like by the time they are done listening to $B$. Such a scheme could work well when comparing images, but appears challenging in an audio context.

\subsection{Integer and Multidimensional Ratings}

A straightforward extension to DeepDrummer would be to ask users to give integer (e.g. 1 to 5) ratings to drum loops instead of asking for a binary \emph{like} or \emph{dislike} classification. Importantly, this would at least allow users to quantify when they are indifferent to certain drum loops. This seems sensible, but we have not investigated this approach, as a binary classification seemed to work well enough for our proof of concept. We have also had internal discussions about the possibility of a multidimensional rating where the user would inform DeepDrummer about the reason why a given drum loop was disliked. For instance, whether a loop was disliked because certain sounds were annoying, or because of the rhythmic component.

\subsection{Four-on-the-floor Constraint}

Earlier version of DeepDrummer had an optional four-on-the-floor constraint. This made it possible to force patterns to include a kick drum at regular intervals. We found that this made DeepDrummer converge more quickly to recognizable drum beats. We chose to not to make use of this constraint in our experiment as we thought that this would better showcase the model's ability to learn its own rhythmic elements, without prior assumptions.

\subsection{Retraining Models}

The current version of DeepDrummer performs interactive learning with a human in the loop. That is, for every new rating provided, the current model is incrementally trained with the new data. An alternative we have tried is to retrain a new model from scratch for every new rating instead, which is not very expensive considering that our datasets are very small. We have found that in practice, interactive learning is much more sample-efficient. We believe this is because, in an interactive learning scheme, new drum loops being generated correspond to the current ``beliefs'' of the latest model. That is, if the model believes that a new drum loop will be \emph{liked} by the user, but the user rates it as \emph{dislike} instead, the training process serves to correct an error in the model. However, if we are retraining new models for every new rating, the models are not necessarily informed by the errors of prior models.

\subsection{Critic Model Design}

In the design of the critic, three main alternatives were considered. The first used a stack of 1D convolutions operating on raw audio waveform data. The second used an FFT transform followed by a stack of 2D convolutions, and the third used an MFCC transform followed by a stack of 2D convolutions. We have found in qualitative evaluations that all of these models appeared to work, but the MFCC-based model produced better results more quickly. We believe this choice is sensible considering that our dataset is very small, which makes it difficult to train 1D convolutions to learn meaningful features from raw audio.

\section{Generation by Metropolis-Hastings Sampling}\label{sec:generation-metropolis-hastings}

Metropolis-Hastings is a standard method for sampling from an unnormalized distribution. See \cite{MacKay2002} for more details.

As mentioned in Section \ref{sec:training-the-critic}, assuming that our critic is minimizing our loss properly, we have that
\begin{equation*}
    f(x) = P\left( \textrm{user rating = like} \hspace{0.2cm} | \hspace{0.2cm} \textrm{input = } x \right).
\end{equation*}
According to Bayes' Rule, we have 
\begin{eqnarray}
    P\left( x \hspace{0.1cm} | \hspace{0.1cm} \textrm{like} \right) & \propto &  P\left( \textrm{like} \hspace{0.1cm} | \hspace{0.1cm} x \right) P(x) \\
    & = & f(x) P(x). \label{eqn:bayes-px}
\end{eqnarray}

Here $P(x)$ is the prior distribution on drum loops without any connection to our users. Note that the generator component in DeepDrummer has a uniform prior probability to generate any drum loop $x$, meaning the $P(x)$ is a constant term in Equation (\ref{eqn:bayes-px}).

To sample from $P\left( x \hspace{0.1cm} | \hspace{0.1cm} \textrm{like} \right)$ with Metropolis-Hastings, we construct a Markov chain $\left\{x^{(t)}\right\}_{t=0}^\infty$ using a proposal distribution $q(x'|x^{(t)})$, where $x'$ is our candidate based on our current $x^{(t)}$. We pick a symmetrical proposal whereby we perturb elements of a pattern $x$ independently and with equal probability. This allows us to cross out the terms from the acceptance ratio:
\begin{equation}
    \alpha = \frac{P\left( x' \hspace{0.1cm} | \hspace{0.1cm} \textrm{like} \right)}{P\left( x \hspace{0.1cm} | \hspace{0.1cm} \textrm{like} \right)} \frac{q\left( x^{(t)} | x'\right)}{q\left( x' | x^{(t)}\right)} = \frac{f(x')}{f(x)}.
\end{equation}
Every time we propose a transition, we accept with probability $\alpha$ and set $x^{(t+1)}=x'$, or we reject otherwise and keep $x^{(t+1)}=x^{(t)}$. When $\alpha \geq 1$ we automatically accept.
This Markov chain will then have $P\left( x \hspace{0.1cm} | \hspace{0.1cm} \textrm{like} \right)$ as stationary distribution.

We can apply a temperature parameter $s>0$ by using $\alpha^{1/s}$ as acceptance ratio instead, with $s\rightarrow0$ leading to a pure optimization problem and $s\rightarrow\infty$ to a scenario where all $x$ are equally desirable.

Concretely, both hill-climbing and Metropolis-Hastings run drum loop candidates through the critic network repeatedly while searching for better candidates, though only the latter is willing to accept worse candidates temporarily in order to escape local maxima.

%
%
%

\section{Hyperparameter Optimization}
\label{sec:hparams-and-mechanical-ear}

We have used Orion \cite{xavier_bouthillier_2019_3478593} extensively to help us make decisions about the model architecture used and other training hyperparameters. This proved to be an invaluable tool.

However, DeepDrummer does not fit the usual profile of deep learning experiment in which static data is being loaded from a hard drive, and where the complete dataset is known ahead of time. In fact, the interactive nature of DeepDrummer means that the training data depends on the current model being trained.

Hyperparameter optimization is generally predicated on the assumption that it is possible to run a large quantity of experiments to find the best set of hyperparameters. This requirement fails when we have a human in the loop.

There were many sanity checks that we could afford to run using static data as training set. For example, it is still possible to do the tradition train/valid/test split to evaluate the potential for a model architecture to generalize well. During test runs with DeepDrummer, we collected as a byproduct a large collection of generated drum loops that we labeled ourselves. These can be used for this purpose, but this is not a guarantee about a model's performance in a different setting.

To test the interactive system, which is fundamentally about a human in the loop, but without requiring a human to be present, we have instead implemented a way to run the interactive system with a neural network serving as proxy for the human. That is, the neural network proxy was trained on a large corpus of music (or previously-labeled drum loops), and we then ran the whole interactive pipeline of DeepDrummer using the proxy instead of the human. DeepDrummer would try to learn from the ``tastes'' of the proxy unaware that it was not a human user.

There are certain limitations to such a setup due to the fact that the proxy will not have rich musical tastes like humans, but it can nevertheless be used to tweak hyperparameters for the whole of DeepDrummer. We deployed Orion again to optimize those hyperparameters. We naturally made sure that we went through the process ourselves afterwards (many times) in order to validate the hyperparameters selected.

\section{Web splash page}
\label{sec:web-splash-page}

\begin{figure}[htb!]
\centerline{
    \includegraphics[width=\columnwidth]{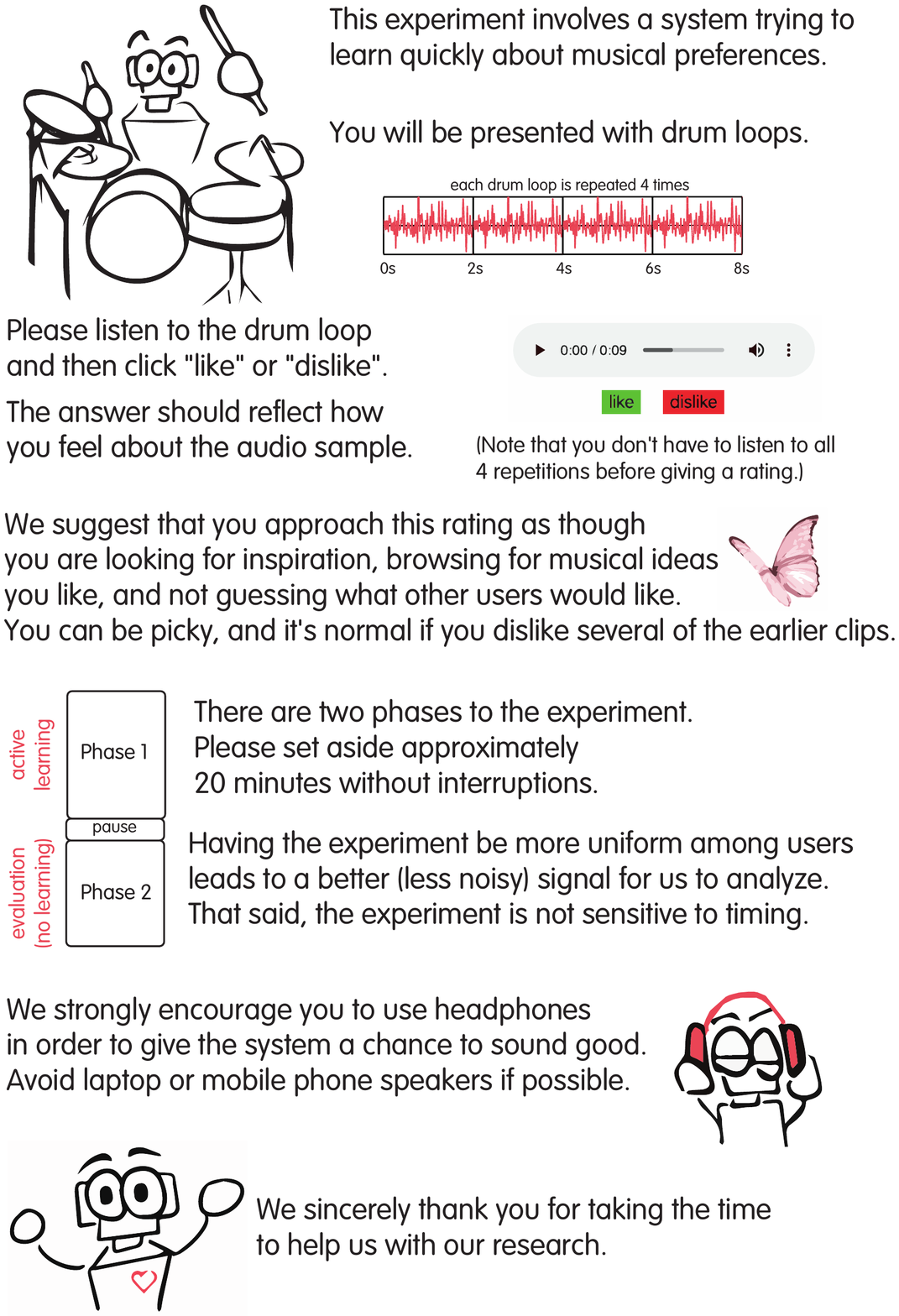}}
    \caption{DeepDrummer splash page for online experiments.
}
\label{fig:web-splash-page}
\end{figure}

We ran experiments our experiments online, giving each participant the same instructions about how to proceed. We show in \figref{fig:web-splash-page} the landing page of DeepDrummer.

\end{document}